\documentclass[letterpaper, 10 pt, conference]{ieeeconf}  % Comment this line out if you need a4paper

\IEEEoverridecommandlockouts                              % This command is only needed if 
\usepackage{authblk}
\usepackage{graphics} % for pdf, bitmapped graphics files
\usepackage{epsfig} % for postscript graphics files
\usepackage{mathptmx} % assumes new font selection scheme installed
\usepackage{times} % assumes new font selection scheme installed
\usepackage{amsmath} % assumes amsmath package installed
\usepackage{amssymb}  % assumes amsmath package installed
\usepackage{cite}
\usepackage{xcolor}
\usepackage{booktabs}
\usepackage{float}
\usepackage{multirow}
\usepackage{stfloats}
\usepackage{caption}
\usepackage{newtxtext, newtxmath}
\usepackage{tabularx}
\usepackage{longtable}
\usepackage{hyperref}

\title{\LARGE \bf
Robust Robot Walker: Learning Agile Locomotion over Tiny Traps
}

\author{Shaoting Zhu$^{1,2}$, Runhan Huang$^{1,2}$, Linzhan Mou$^{2,3}$, Hang Zhao†$^{1,2}$% <-this % stops a space
\thanks{$^{1}$IIIS, Tsinghua University, Beijing, China}%
\thanks{$^{2}$Shanghai Qi Zhi Institute, Shanghai, China}%
\thanks{$^{3}$GRASP Lab, University of Pennsylvania, Philadelphia, PA, USA}%
\thanks{† Corresponding author. E-mail: \tt\small zhaohang0124@gmail.com}%
}

\begin{document}

\maketitle
\thispagestyle{empty}
\pagestyle{empty}

\begin{figure}[h]
    \centering
    \includegraphics[width=0.9\linewidth]{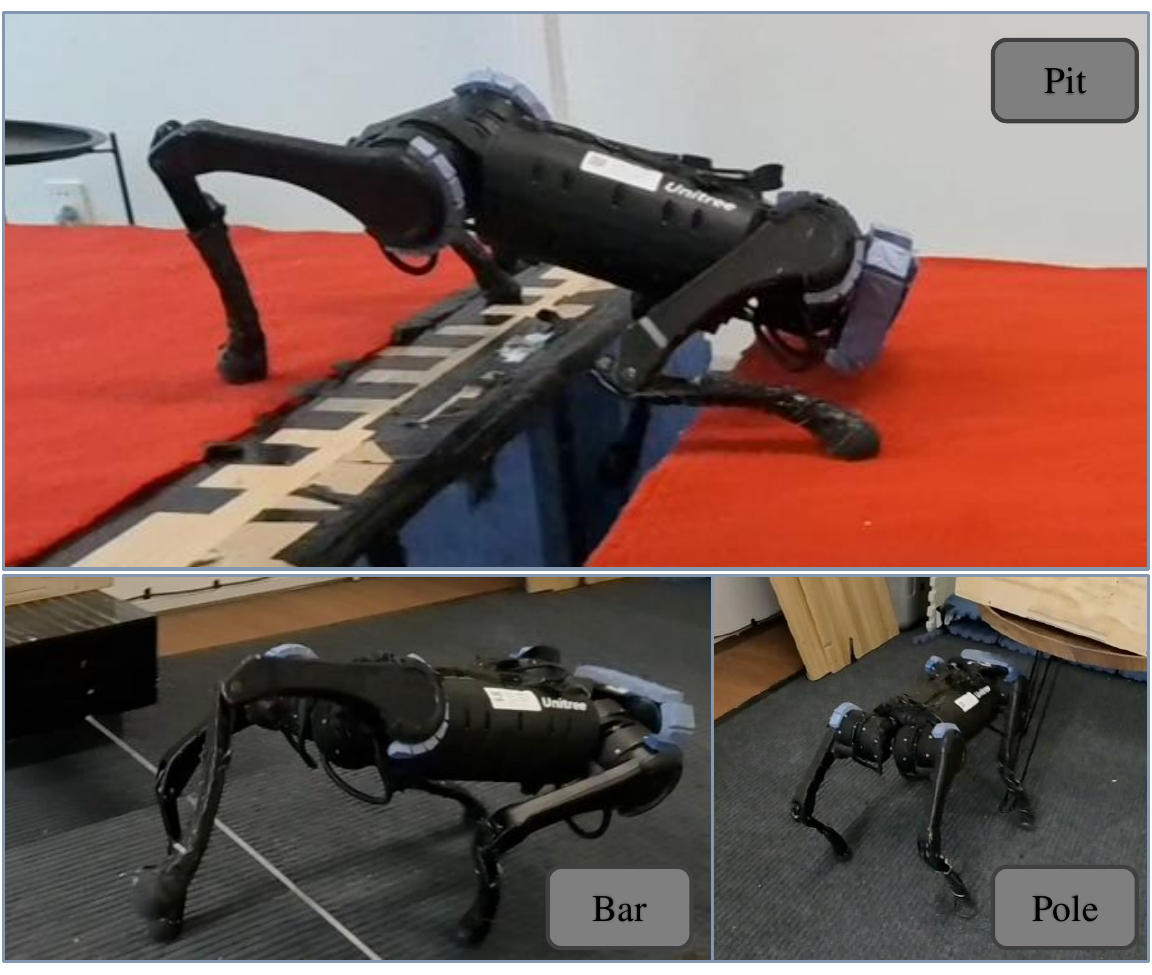}
    \label{fig:teaser}
    \vspace{-1mm}
    \caption{Our \texttt{Robust Robot Walker} is passing through various challenging ``tiny traps'' including Pit, Bar, and Pole solely relying on its proprioception.}
\end{figure}

\vspace{-4mm}

%%%%%%%%%%%%%%%%%%%%%%%%%%%%%%%%%%%%%%%%%%%%%%%%%%%%%%%%%%%%%%%%%%%%%%%%%%%%%%%%
\begin{abstract}

Quadruped robots must exhibit robust walking capabilities in practical applications. In this work, we propose a novel approach that enables quadruped robots to pass various small obstacles, or ``tiny traps''. Existing methods often rely on exteroceptive sensors, which can be unreliable for detecting such tiny traps. To overcome this limitation, our approach focuses solely on proprioceptive inputs. We introduce a two-stage training framework incorporating a contact encoder and a classification head to learn implicit representations of different traps.
Additionally, we design a set of tailored reward functions to improve both the stability of training and the ease of deployment for goal-tracking tasks. To benefit further research, we design a new benchmark for tiny trap task. Extensive experiments in both simulation and real-world settings demonstrate the effectiveness and robustness of our method. 
Project Page: \href{https://robust-robot-walker.github.io/}{https://robust-robot-walker.github.io/}.

\end{abstract}

%%%%%%%%%%%%%%%%%%%%%%%%%%%%%%%%%%%%%%%%%%%%%%%%%%%%%%%%%%%%%%%%%%%%%%%%%%%%%%%%
\section{INTRODUCTION}

\begin{figure}[t]
    \centering
    \includegraphics[width=\linewidth]{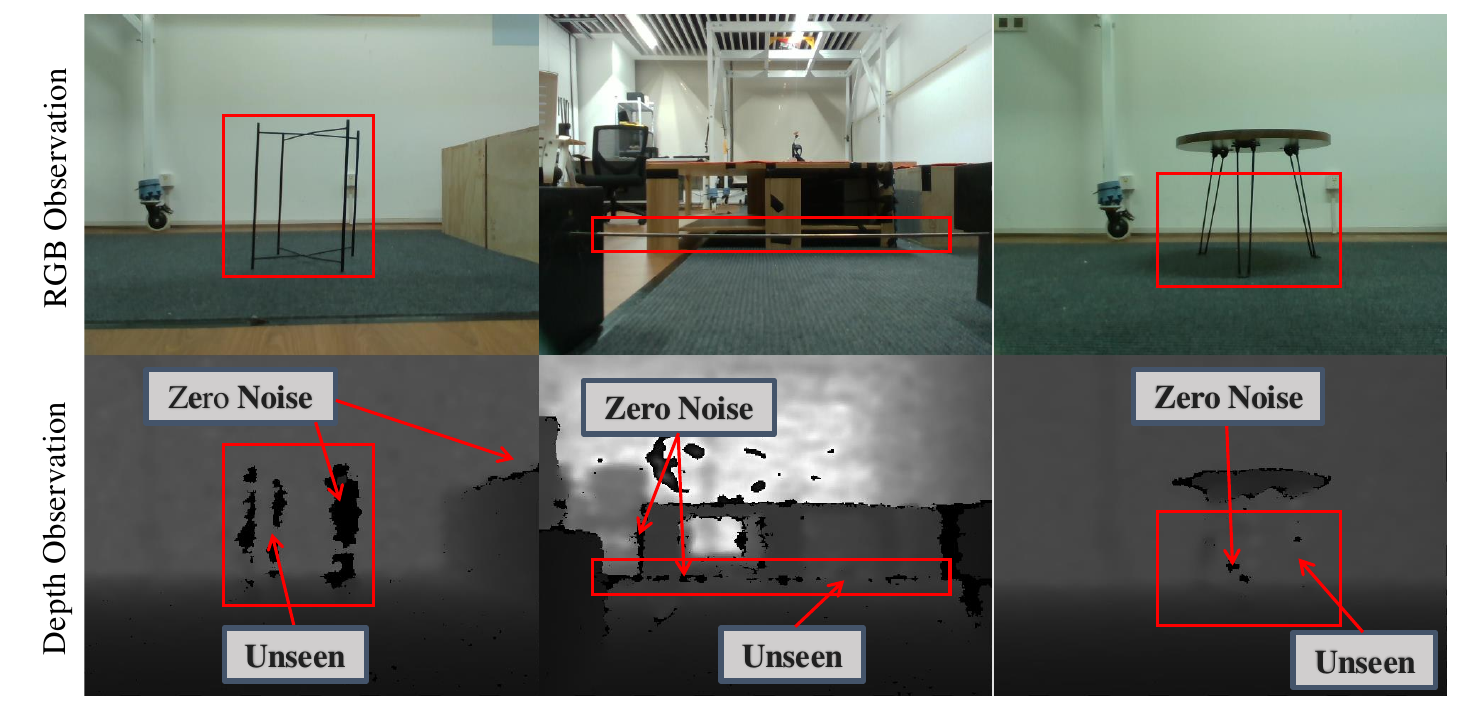}
    \vspace{-6mm}
    \caption{Camera unreliability in fine-grained trap scenarios.
    }
    \label{fig:depth_show}
    \vspace{-6.5mm}
\end{figure}

Humans and animals have the ability to walk robustly in complex environments, relying on proprioception to avoid various obstacles such as strings, poles, and ground pits. However, these same challenges present significant difficulties for robots. In real-world scenarios, seemingly minor traps can severely impact a robot’s mobility. Many of these obstacles are tiny or positioned below or behind the robot, making them difficult to detect with external sensory devices like depth cameras, as shown in Fig.~\ref{fig:depth_show}. Small objects like narrow bars or poles often produce unreliable data in depth images, appearing intermittently noisy or as a dense patch at zero distance, making them indistinguishable from the edge noise of other obstacles. Additionally, since realistic RGB images cannot be accurately rendered in simulations, using them in real-world applications is limited due to a significant sim-to-real gap. This arises the need for developing control policies that enable robots to overcome such trap-type obstacles without relying on additional sensory equipment.

% \vspace{-3mm}
% \vspace{-4mm}

Learning agile locomotion over tiny traps presents several challenges. First, frameworks~\cite{agarwal2023legged, zhuang2023robot} that rely on exteroceptive inputs are ineffective for tasks involving tiny traps, as both RGB and depth images are difficult to leverage in these scenarios. Second, with incomplete perceptual information, it is difficult to learn a blind walking policy directly from scratch. Some privileged information is required to guide the training. Lastly, while some goal-tracking frameworks~\cite{rudin2022advanced, zhang2023learning} have addressed the above issues, they often lack omnidirectional movement capabilities or rely heavily on external localization techniques. Moreover, these frameworks frequently employ sparse rewards, which cause the instability of training and complicate real-world deployment.

To address the challenges of learning agile locomotion over tiny traps using proprioception, we propose a novel solution with several key contributions. 
\vspace{-0.5mm}
\begin{itemize}
    \item First, we introduce a \textbf{two-stage training framework} that \textbf{relies solely on proprioception}, enabling a robust policy that successfully passes tiny traps in both simulation and real-world environments.
    \item Second, we develop an \textbf{explicit-implicit dual-state estimation paradigm}, utilizing a contact encoder to estimate contact forces on different robot links and a classification head to enhance the learning of contact representations.
    \item Third, we \textbf{redefine the task as goal tracking}, rather than velocity tracking, and incorporate carefully designed \textbf{dense reward functions and fake goal commands}. This approach achieves approximate omnidirectional movement without motion capture or additional localization techniques in real-world, significantly improving training stability and adaptability across environments.
    \item Finally, we introduce \textbf{a new benchmark for tiny trap tasks} and conduct extensive experiments in both simulation and real-world scenarios, demonstrating the robustness and effectiveness of our method.
    % These experiments, alongside analyses such as t-SNE visualizations and saliency maps, 
    
\end{itemize}   

\vspace{-2mm}
\begin{figure*}[htbp]
    \centering
    \includegraphics[width=0.95\textwidth]{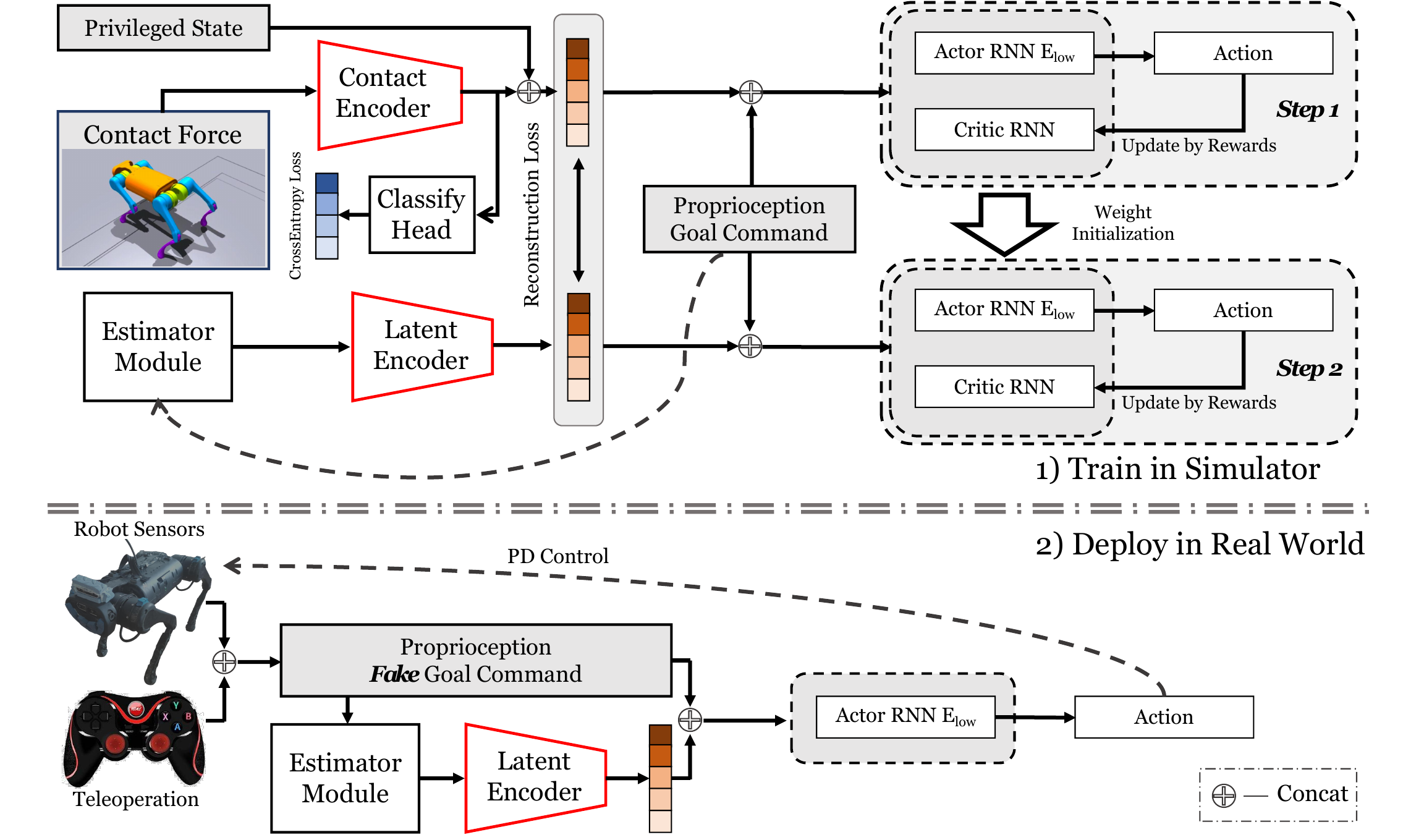}
    \caption{The training and deployment overview of \texttt{Robust Robot Walker}. Our method achieves explicit-implicit dual-state estimation and approximate omnidirectional movement. Each color on the quadruped robot corresponds to a type of joint link: Orange-Base, Yellow-Hip, Blue-Thigh, Purple-Calf, and Foot (Unseen).}
    \label{fig:pipeline}
    \vspace{-7mm}
\end{figure*}

% In short, we summarize our contributions as follows:
% \textbf{1)} We propose a two-stage training framework and train a policy using only proprioception. The policy is successfully deployed on the real robot. It can effectively pass through several tiny traps in both simulation and the real world.
% \textbf{2)} We introduce an explicit-implicit dual-state estimation diagram. We use a contact encoder to estimate the discrete contact force of different links on the robot. Besides, a classification head is used to learn the latent contact better.
% \textbf{3)} We define the task as goal tracking instead of velocity tracking. With carefully designed dense reward functions and fake goal commands, we achieve approximate omnidirectional movements without motion capture and any other localization techniques. This greatly improves the training stability and the environment adaptability.
% \textbf{4)} We introduce a brand new tiny trap benchmark and conduct lots of experiments in both simulation and the real world. In addition, we further analyze different fake commands, t-SNE visualization of latent space, and saliency maps. All the experiments fully demonstrate the robustness and effectiveness of our method.

\section{RELATED WORKS}
\vspace{-0.5mm}
\subsection{Learning-Based Locomotion}
\vspace{-0.5mm}
The deep reinforcement learning (DRL) paradigm has demonstrated its ability to learn legged locomotion behaviors in simulators\cite{makoviychuk2021isaac}. Unlike model-based control methods\cite{horvat2017model}, DRL enables robots to handle corner cases in simulation through an end-to-end learning process, resulting in robust, transferable policies. These simulation-trained policies allow robots not only to walk on flat terrain\cite{hwangbo2019learning}, but also to achieve high-speed running\cite{ji2022concurrent,margolis2024rapid}, traverse muddy or uneven terrains\cite{kumar2021rma}, stand on rear legs\cite{su2024leveraging, smith2023learning}, open doors\cite{su2024leveraging}, climb stairs\cite{wu2023learning, lee2020learning}, scale rocky terrain\cite{cheng2024quadruped,zhang2023learning}, and even perform high-speed parkour\cite{rudin2022advanced,zhuang2023robot, cheng2024extreme, hoeller2024anymal, luo2024pie}.

\vspace{-1.5mm}

\subsection{Collision Detection}
\vspace{-0.5mm}
Collisions in robots can be classified into several stages: the pre-collision phase~\cite{haddadin2017robot, sisbot2012human, ebert2002safe}, collision detection phase~\cite{haddadin2017robot, takakura1989approach, morinaga2003collision}, collision isolation phase~\cite{haddadin2017robot, vorndamme2017collision}, collision identification phase~\cite{haddadin2017robot, vorndamme2017collision, kuntze2003fault}, collision classification phase~\cite{golz2015using}, collision reaction phase~\cite{haddadin2008role, haddadin2008collision, parusel2011modular}, and post-collision phase~\cite{parusel2011modular, golz2015using}. Collision detection methodologies for quadruped robots are generally divided into model-based and model-free approaches. Model-based methods typically employ state estimation techniques~\cite{van2022collision, hwangbo2016probabilistic, camurri2017probabilistic}. Some approaches leverage exteroceptive sensors~\cite{maravgakis2023probabilistic}, while others rely purely on proprioception for estimation~\cite{barasuol2019detection}. Model-free methods, on the other hand, involve training neural network-based contact estimators through deep reinforcement learning, which can be either implicit~\cite{nahrendra2023dreamwaq} or explicit~\cite{cheng2024quadruped}.

% \vspace{-1mm}

% \subsection{Navigation command}
% Using velocity command to train the policy on challenging terrains, the robot may learn to bypass the obstacle instead of crossing it, or even rebouncing back and forth. ~\cite{cheng2024quadruped} proposed to use elaborated rewards to prevent robots from cheating in this way. To solve the problem, researchers proposed to use the navigation command~\cite{zhang2023learning, rudin2022advanced}, which encourages the robot to reach a given target within a given time, but relies on localization techniques and suffers from sparse rewards.

\vspace{-0.5mm}

\section{METHOD}

\subsection{Task Definition}
\begin{figure}[h]
    \centering
    \vspace{-3mm}
    \includegraphics[width=0.8\linewidth]{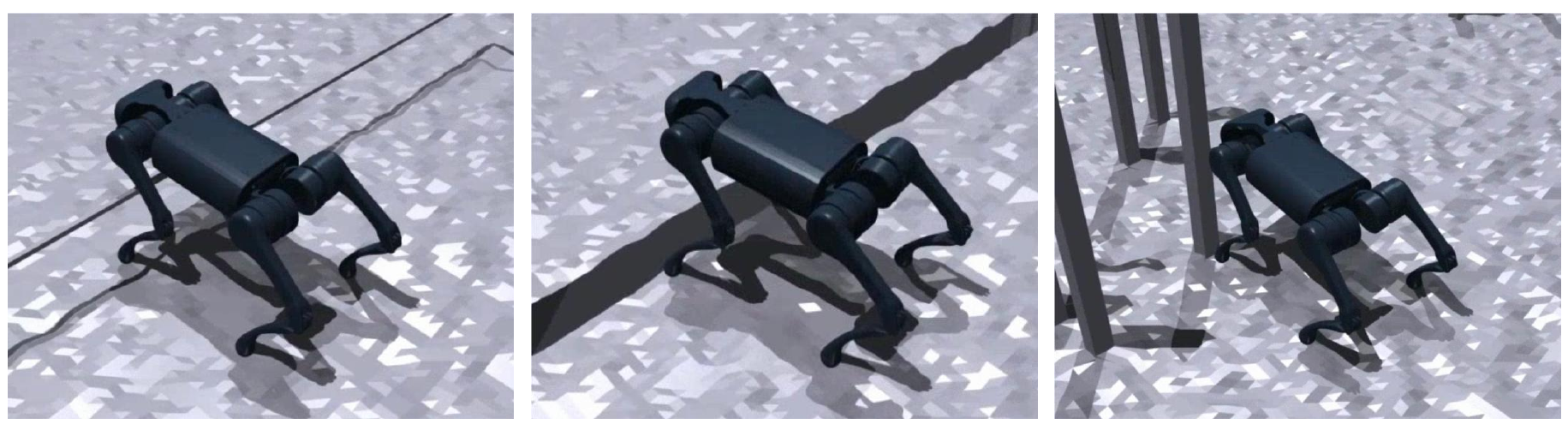}
    \vspace{-2mm}
    \caption{Three categories of tiny traps: Bar, Pit, and Pole.}
    \label{fig:show_trap}
    \vspace{-4mm}
\end{figure}

The task is defined as passing through a set of tiny traps, denoted as $\mathcal{T}$. An onboard camera typically fails to detect such traps. These traps are categorized into three main types: \textit{Bar}, \textit{Pit}, and \textit{Pole}, as illustrated in Fig.~\ref{fig:show_trap}. A \textit{Bar} refers to a horizontal thin bar positioned over a plane at a height below the quadruped robot's head. A \textit{Pit} is a small depression between two planes that can cause the robot's legs to slip, and it is not visible from the normal front view. A \textit{Pole} is a thin, upright pole standing on a plane. When encountering these traps, the robot is provided with \texttt{constant} control commands and must adjust its speed autonomously to pass them. It's important to note that in training and demonstrations, the traps include both thin and thick \textit{Bars}/\textit{Poles}. While the onboard camera can detect thicker ones, they are included only to showcase the generalization capabilities of our control policy.

% The task is defined as passing through tiny traps $\mathcal{T}$. An onboard camera usually cannot perceive a tiny trap $\mathcal{T}$ as shown in Fig.~\ref{fig:show_trap}. There are mainly three categories: \textit{Bar}, \textit{Pit}, and \textit{Pole}. \textit{Bar} is a horizontal thin bar over a plane with a height lower than the quadruped robot's head. \textit{Pit} is a small pit between two planes that can cause the robot's legs to fall, which cannot be seen by the normal front view. \textit{Pole} is a thin pole that stands upright on a plane. When passing through these traps, the control commands given to the robot are \texttt{constant}. Robot needs to make some speed changes on its own to deal with these traps. Note that in our training and demo, there are both thin and thick \textit{Bar}/\textit{Pole}. The camera can capture the latter, but they are only to demonstrate the generalization of our control policy.

\vspace{-1.5mm}

\subsection{Reinforcement Learning Setting}
\vspace{-0.5mm}

We decompose the locomotion control problem into discrete locomotion dynamics, with a discrete time step $d_t=0.02s$. We use the Proximal Policy Optimization algorithm (PPO)\cite{schulman2017proximal} to optimize our policy. Inspired by \cite{rudin2022advanced, zhang2023learning}, we formulate the problem as goal tracking instead of velocity tracking.

\textbf{State Space:} The entire process includes the following four types of observation: proprioception $\boldsymbol{p}_t$, privileged state $\hat{\boldsymbol{s}}_t$, contact force $\hat{\boldsymbol{c}_t}$, and goal command $\boldsymbol{g}_t$. \textbf{1) Proprioception $\boldsymbol{p}_t$} contains gravity vector and base angular velocity from IMU, joint positions, joint velocities, and last action. \textbf{2) Privileged state $\hat{\boldsymbol{s}}_t$} contains base linear velocity (unreliable from IMU) and the ground friction. \textbf{3) Contact force $\hat{\boldsymbol{c}_t}$} includes contact force with environment meshes of each joint link, refer to the robot link in different colors in Fig.~\ref{fig:pipeline}. Each contact force is clipped, and normalized to $[-1,1]$. \textbf{4) Goal command} $\boldsymbol{g}_t$ includes goal position $\Delta G=(\Delta x, \Delta y, \Delta z)$ relative to the current robot frame, and the time remaining to complete $\Delta t$. At the beginning of one episode, we randomly sample a goal position fixed in the world frame and set $\Delta t$ to the episode length. In every time step, we calculate $\Delta G$ based on the robot pose, and update $\Delta t$ to the last time of the current episode. Since our task does not involve height change, $\Delta z$ is always set to zero. To facilitate the training of a standstill state, we set $\Delta G=(0, 0, 0)$ when $\Vert \Delta G\Vert_2<0.2$.

% We use a two-step training process, as illustrated in Fig.~\ref{fig:pipeline}. In the first step, the policy leverages all available observation $[\boldsymbol{p}_t, \hat{\boldsymbol{s}}_t, \hat{\boldsymbol{c}_t}, \boldsymbol{g}_t]$, while in the second step and real-world deployment, it relies only on $[\boldsymbol{p}_t, \boldsymbol{g}_t]$.

\textbf{Action Space:} The action space $\boldsymbol{a}_t\in\mathbb{R}^{12}$ is the desired joint positions of 12 joints.

\vspace{-1.5mm}

\subsection{Reward Function}
\vspace{-0.5mm}
\label{subsec:reward}

The reward function has three components: task reward $r_t^T$, regularization reward $r_t^R$, and style reward $r_t^S$. The total reward is the sum of three items: $r_t = r_t^T + r_t^R + r_t^S$.

\textbf{1) Task reward.} $r_{goal}$, $r_{heading}$, and $r_{finish}$ are the main components. Unlike previous works, we define the $r_{goal}$ as a dense reward throughout the entire episode. 
% In our setting, the robot can only perceive the tiny traps when in contact with it. It cannot predict the traps in advance at a long distance and detour against the reward function, so there is no need for free movement exploration as\cite{zhang2023learning}. In addition, this greatly improves the stability of training.
In our setup, the robot can only perceive tiny traps upon contact, meaning it cannot predict and avoid them from a distance against the reward function. This eliminates the need for free movement exploration as seen in~\cite{zhang2023learning}, and significantly enhances the stability of the training process.
\begin{equation}
    r_{goal}=\frac{1}{0.4+\Vert \Delta G\Vert_2},
\end{equation}

We aim for the robot to always move toward the goal. Without the heading reward, it may encounter traps horizontally or backward, which is equivalent to multi-task learning, and greatly increases the difficulty of training. Although heading reward allows the robot to only pass through the trap in a straight or with a small angle, it greatly speeds up convergence and improves the passing success rate. $\epsilon=10^{-6}$.
\vspace{-1mm}
\begin{equation}
r_{heading}=
    \left \{ \begin{matrix}
    \vspace{2mm}
    \displaystyle \frac{\Delta x+\epsilon}{\Vert \Delta G\Vert_2+\epsilon}, &\Vert \Delta G\Vert_2 \neq 0\\
    1,&\Vert \Delta G\Vert_2 = 0\\
    \end{matrix} \right. ,
\end{equation}

Besides, the policy should be able to operate stably for a long time on the real robot. Therefore, we introduce a finish reward to encourage the robot to stand still when near the goal. When $\Delta G=(0, 0, 0)$:
\vspace{-1mm}
\begin{equation}
    r_{finish\_pos}=\sum^{12}_{i=1}{{\lvert q-q_{default}\rvert}},\\
\end{equation}
\vspace{-6mm}
\begin{equation}
    r_{finish\_vel}=\Vert v\Vert+\Vert \omega\Vert,
\end{equation}
\vspace{-5mm}
\begin{equation}
    r_{finish}=\lambda_1\cdot r_{finish\_pos}+\lambda_2\cdot r_{finish\_vel}.
\end{equation}
\vspace{-6mm}

\textbf{2) Regularization reward.} Regularization reward is designed to make the robot move smoothly, safely, and naturally. A key component of this is the velocity limit reward. Unlike previous work, our goal is not for the robot to reach the speed limit, which is unsafe in a real deployment.
\vspace{-2mm}
\begin{equation}
    r_{vel\_limit}=(\omega_{z}<\omega_{limit})\cdot(\Vert v_{x,y}\Vert_2<v_{limit})
    \vspace{-2mm}
\end{equation}

Other regularization rewards include stall, leg energy, dof vel, etc. The details can be found in \textit{Appendix A}.

\textbf{3) Style reward.} We use the Adversarial Motion Priors (AMP) style reward to gain a natural gait and speed up convergence following \cite{peng2021amp, wu2023learning}.

\begin{figure*}[t]
    \centering
    \makebox[\textwidth][c]{\includegraphics[width=1.0\textwidth]{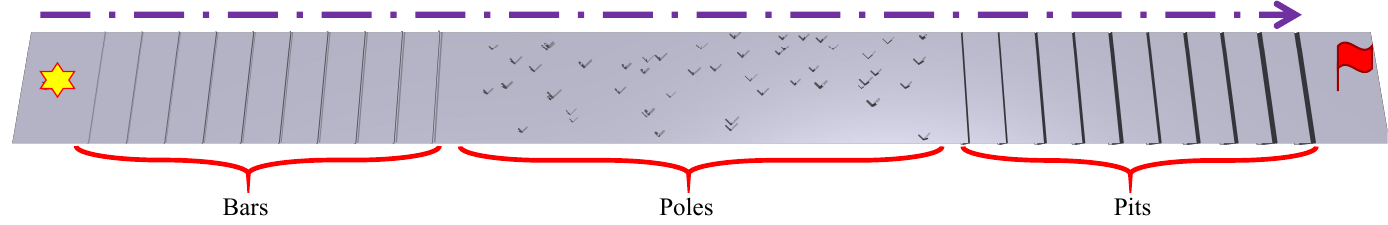}}  
    \vspace{-6mm}
    \caption{\textbf{Tiny Trap Benchmark.} Robots begin on the left side and must pass through tiny traps to reach the goal on the right side.}
    \label{fig:benchmark}
    \vspace{-6mm}
\end{figure*}

\subsection{Training and Deployment}
\label{subsec:train_deploy}

\textbf{1) Training.} To effectively learn the privileged state and improve the performance, we employ a two-stage Probability Annealing Selection (PAS) framework~\cite{zhu2024cross} instead of the traditional teacher-student imitation learning approach, as illustrated in Fig.~\ref{fig:pipeline}. Further details are in \textit{Appendix B}.

In the first step of training, the policy can access all the information $[\boldsymbol{p}_t, \hat{\boldsymbol{s}}_t, \hat{\boldsymbol{c}_t}, \boldsymbol{g}_t]$ as observation. In addition, we use explicit-implicit dual-state learning similar to \cite{luo2024pie}. The contact force $\hat{\boldsymbol{c}_t}$ is first encoded by a contact encoder to an implicit latent, and concatenated with explicit privileged state $\hat{\boldsymbol{s}}_t$ to the dual-state $\hat{\boldsymbol{l}_t}=[ Enc(\hat{\boldsymbol{c}_t}), \hat{\boldsymbol{s}}_t]$. In addition, we introduce a classification head to guide the policy in learning the connection between the contact force distribution and the trap category. In fact, the category of trap can be seen as a very strong explicit privileged state. We use cross-entropy loss as the classification loss. Previous work \cite{cheng2024quadruped} uses boolean values as the collision state for prediction. We find this will cause the sim-to-real problem, referring to the experiment Sec.~\ref{subsec:real_exp}. Besides, we pre-train the estimator module in the first step using L2 loss to reconstruct $\hat{\boldsymbol{l}_t}$. In summary, the total optimization target is:
\vspace{-1mm}
\begin{equation}
    L_{surrogate}+L_{value}+L_{recon}+L_{classify}+L_{discriminator}.
    \vspace{-1mm}
\end{equation}

In the second step of training, the policy can only access [$\boldsymbol{p}_t$, $\boldsymbol{g}_t$] as observation. We initialize the weight of the estimator and the low-level RNN copied from the first training step. Probability Annealing Selection is used to gradually adapt policies to inaccurate estimates while reducing the degradation of the Oracle policy performance.

\vspace{-5mm}
\begin{align}
\boldsymbol{l}_t &= \text{Estimator}(\boldsymbol{p}_t,\boldsymbol{g}_t),\\
\boldsymbol{i}_t &= \text{Probability Selection~}(\boldsymbol{P}_t, \boldsymbol{l}_t, \hat{\boldsymbol{l}_t}),\\
\boldsymbol{a}_t &= \text{Actor}~E_{low}(\boldsymbol{i}_t, \boldsymbol{p}_t,\boldsymbol{g}_t),\\
\text{Probability~} \boldsymbol{P}_t &= \boldsymbol{\alpha}^{iteration}.
\end{align}
\vspace{-6mm}

To enhance training stability, we increase the batch size by using 4,096 parallel robots, each performing 50 steps. Moreover, the episode length is reduced from 20 seconds to 8 seconds.
% We terminate the episode when the robot's base tilts too much sideways, going beyond 0.8rad around its forward axis, or too much up and down, exceeding 1.0rad around its vertical axis. Also, if the robot doesn't move much for more than a second, the episode stops. To improve the stability of the training process, we increase the batch size by using 4096 parallel robots performing 50 steps. Additionally, we reduce the episode length from 20s to 8s.
More details of dynamic randomization and trap terrain curriculum are in the \textit{Appendix C} and \textit{Appendix D}.

\textbf{2) Deployment.} 
% Unlike previous works, our policy can achieve approximate omnidirectional movement ability by teleoperation without motion capture and other auxiliary localization techniques. As shown in Fig.~\ref{fig:pipeline}, the dual-state $\boldsymbol{l}_t$ is predicted by the Estimator Module and Latent Encoder, then sent to low-level Actor RNN combined with proprioception from robot sensors. We use a teleoperator to obtain fake goal commands. It is composed of constant values of the $\Delta G$ and $\Delta t$.
Unlike previous works, our policy achieves approximate omnidirectional movement by teleoperation without motion capture or other auxiliary localization techniques. As shown in Fig.~\ref{fig:pipeline}, the dual-state $\boldsymbol{l}_t$ is predicted by the Estimator Module and Latent Encoder, then passed to the low-level Actor RNN combined with proprioceptive from robot sensors. A teleoperator is used to generate fake goal commands, consisting of constant values for $\Delta G$ and $\Delta t$.

% Under the well-designed task reward function and their proportions, the policy learns different movement strategies for goal positions at different distances. Specifically, when the goal position is far away, the robot preferentially rotates in place toward the goal and then moves forward; When the goal position is at a medium distance, the robot moves towards the goal while turning; When the target point is very close, the robot does not turn and moves directly towards the target point. This is because turning towards the goal will increase $r_{heading}$, but it takes extra time to decrease $r_{goal}$. These two rewards reach a dynamic balance in the training process to get the maximum reward. Taking advantage of this feature, we can use different combinations of goal directions and distances to obtain different movements.
With the well-designed task reward function and their proportions, the policy learns different movement strategies based on the distance to the goal. For large distance, the robot first rotates in place to face the goal, then moves forward. For medium distance, the robot simultaneously moves toward the goal while turning. For small distance, the robot moves directly to it without turning. This behavior arises from the interplay between $r_{heading}$, which increases when the robot turns toward the goal, and $r_{goal}$, which decreases with time spent turning. During training, these two rewards reach a dynamic balance, maximizing overall reward. By leveraging this feature, we can achieve different movement patterns using various combinations of goal directions and distances.

% When $\Vert \Delta G\Vert_2$ is a little more than stop threshold 0.2, the angular velocity of the robot is approximately zero. This enables the translational movement of the robot, including the most commonly used forward, backward, left, and right translations. When $\Vert \Delta G\Vert_2$ is much more than stop threshold 0.2, the robot will have a large angular velocity. For example, when the goal is constantly 1.0 meters to the right or left of the robot, the robot will keep turning right or left in place. Using these two tricks, we can already complete the basic operation. More combinations of velocities can be found in Sec.~\ref{subsec:additional_exp}. Although the robot has not been trained with constant $\Delta G$ and $\Delta t$, it can zero-shot for deployment, and there is little performance degradation. In addition, the robot can complete a variety of movements one-by-one by repeatedly changing $\Delta G$, without resetting the policy and the robot. With standstill training, we can start the robot from a stationary state ($\Vert \Delta G\Vert_2=0$), and then calmly complete various operations.
When $\Vert \Delta G\Vert_2$ is just above the stop threshold of 0.2, the robot’s angular velocity is nearly zero, enabling translational movements such as forward, backward, left, and right. When $\Vert \Delta G\Vert_2=1.0$ to the right or left, the robot will continuously turn right or left in place. By leveraging these two tricks, we can accomplish basic operations. Additional combinations of velocities are discussed in Sec.~\ref{subsec:additional_exp}. Although the robot has not been explicitly trained with constant $\Delta G$ and $\Delta t$, it can zero-shot deploy with no performance degradation. Furthermore, the robot can perform a variety of movements sequentially by changing $\Delta G$ and $\Delta t$, without needing to reset the policy or the robot. With standstill state, the robot can start from a stationary state ($\Vert \Delta G\Vert_2=0$) and seamlessly execute operations.

% In practice, the left side of the joystick is used to control the direction of the goal command (i.e. the ratio of $\Delta x$ and $\Delta y$), and the right side is used to control the distance $\Vert \Delta G\Vert_2$ of the goal. In addition, some keys on the teleoperator can toggle the value of $\Delta t$.
In practice, the left side of the joystick controls the direction of the goal command (i.e., the ratio of $\Delta x$ to $\Delta y$), while the right side controls the distance $\Vert \Delta G\Vert_2$ to the goal. Additionally, some keys on the teleoperator can toggle the value of $\Delta t$.
\vspace{-1mm}
\section{EXPERIMENTS}
\vspace{-1mm}
\subsection{Experiment Setup}
\vspace{-0.5mm}
We use the IsaacGym\cite{makoviychuk2021isaac} for policy training and deploy 4,096 quadruped robot agents on a single NVIDIA RTX 3090. We first train 12,000 iterations on plane terrain, then train 30,000 iterations for both stages on trap terrain. The control policy within both the simulator and the real world operates at a frequency of 50 Hz. We deploy our policy on the Unitree A1 quadruped robot which has an NVIDIA Jetson Xavier NX as the onboard computer. In the real-world deployment, the robot receives the goal command from the teleoperator through ROS message and runs the low-level control policy to predict desired joint positions for PD control $(K_p=40, K_d=0.5)$.

\begin{table*}[htbp]
    \centering
    \caption{The comparison results with other locomotion policies in our benchmark.}
    \vspace{-1mm}
    \label{tab:comparison}
    \setlength\tabcolsep{8pt}
    \fontsize{8}{10}\selectfont
    \begin{tabular}{c|cccc|cccc|cccc}
        \toprule
        \multirow{2}{*}{\textbf{Method}} & \multicolumn{4}{c|}{\textbf{Success Rate} $\uparrow$}&  \multicolumn{4}{c|}{\textbf{Average Pass Time (s)} $\downarrow$}&  \multicolumn{4}{c}{\textbf{Average Travel Distance (m)} $\uparrow$}\\
         & Mix& Bar& Pit& Pole& Mix& Bar& Pit& Pole& Mix& Bar& Pit& Pole\\
         \midrule
         \textbf{\textbf{Ours}} &  \textbf{0.934}&   \textbf{0.941}&  \textbf{0.902}&  \textbf{0.702}&  \textbf{119.03}& \textbf{130.65}& \textbf{111.91} &\textbf{156.40} & \textbf{56.37} & \textbf{55.54} & \textbf{56.22} & \textbf{44.06}\\
         w/o goal command&  0.000&   0.000&  0.000&  0.279&  300.00& 300.00& 300.00 &254.17& 3.41& 3.81 &11.32 &24.15\\
         Ours w/ Boolean &  \underline{0.779} & \underline{0.718} &  \underline{0.656} &  \underline{0.658} &  \underline{144.66} & \underline{169.59} & \underline{161.17} & \underline{167.21} & \underline{52.40} & \underline{50.12} & \underline{49.28} & \underline{41.65} \\
         RMA &  0.000&   0.000&  0.000&  0.067& 300.00& 300.00& 300.00 &291.99& 5.27& 7.45 &10.21 &17.17\\
         MoB &  0.000&   0.000&  0.000&  0.000&  300.00& 300.00& 300.00 &300.00& 2.90& 4.13 &4.42 &6.57\\
         HIMLoco &  0.000&  0.000& 0.000& 0.000& 300.00& 300.00& 300.00& 300.00& 3.61& 4.16 &16.73& 16.15\\
         \bottomrule
    \end{tabular}
    \vspace{-1mm}
\end{table*}

\begin{table*}[htbp]
    \centering
    \caption{The ablation results of different link. \texttt{B}-base link, \texttt{H}-hip link, \texttt{T}-thigh link, \texttt{C}-calf link, and \texttt{F}-foot link.}
    \vspace{-1mm}
    \label{tab:ablation}
    \setlength\tabcolsep{8pt}
    \fontsize{8}{10}\selectfont
    \begin{tabular}{c|cccc|cccc|cccc}
        \toprule
        \multirow{2}{*}{\textbf{Method}} & \multicolumn{4}{c|}{\textbf{Success Rate} $\uparrow$}&  \multicolumn{4}{c|}{\textbf{Average Pass Time (s)} $\downarrow$}&  \multicolumn{4}{c}{\textbf{Average Travel Distance (m)} $\uparrow$}\\
         & Mix& Bar& Pit& Pole& Mix& Bar& Pit& Pole& Mix& Bar& Pit& Pole\\
         \midrule
         \textbf{\texttt{BHTCF} (Ours)} &  \textbf{0.934}&   \textbf{0.941}& \textbf{0.902}&  0.702&  \textbf{119.03}& 130.65& \textbf{111.91} &\underline{156.40} & \textbf{56.37} & \textbf{55.54} & \textbf{56.22} & \underline{44.06}\\
         \texttt{BTCF}&  \underline{0.918}&   0.925&  0.901&  \underline{0.704}&  122.58& 138.98& \underline{111.99} &158.91& \underline{56.21} & \underline{55.10} & \underline{55.97} & 41.59\\
         \texttt{TCF}&  0.828&   0.815&  0.820&  0.618&  144.47& 163.72& 134.32 &176.16& 53.03 & 48.44 & 52.62 & 37.77\\
         \texttt{TC}&  0.904&   \underline{0.929}&  \textbf{0.902}&  0.687&  \underline{120.88}& \textbf{125.19}& 113.36 &159.21& 55.58 & 54.89 & 54.96 & 41.54\\
         \texttt{BTC}&  0.901&   0.902&  0.899&  \textbf{0.729}&  123.59& 136.90& 112.15 &\textbf{153.41}& 55.70 & 53.18 & 54.17 & \textbf{46.03}\\
         \texttt{T}&  0.864&   0.919&  0.850&  0.692&  128.97& 
         \underline{128.28}& 122.87 &159.60& 53.97 & 53.44 & 53.23 & 41.87\\
         Prop \texttt{(None)}&  0.819&   0.874&  0.796&  0.655&  142.84& 148.01& 139.03 &164.72& 52.10 & 50.02 & 49.85 & 39.40\\
         \bottomrule
    \end{tabular}
    \vspace{-4mm}
\end{table*}

\begin{figure*}[t]
    \centering
    \includegraphics[width=0.85\textwidth]{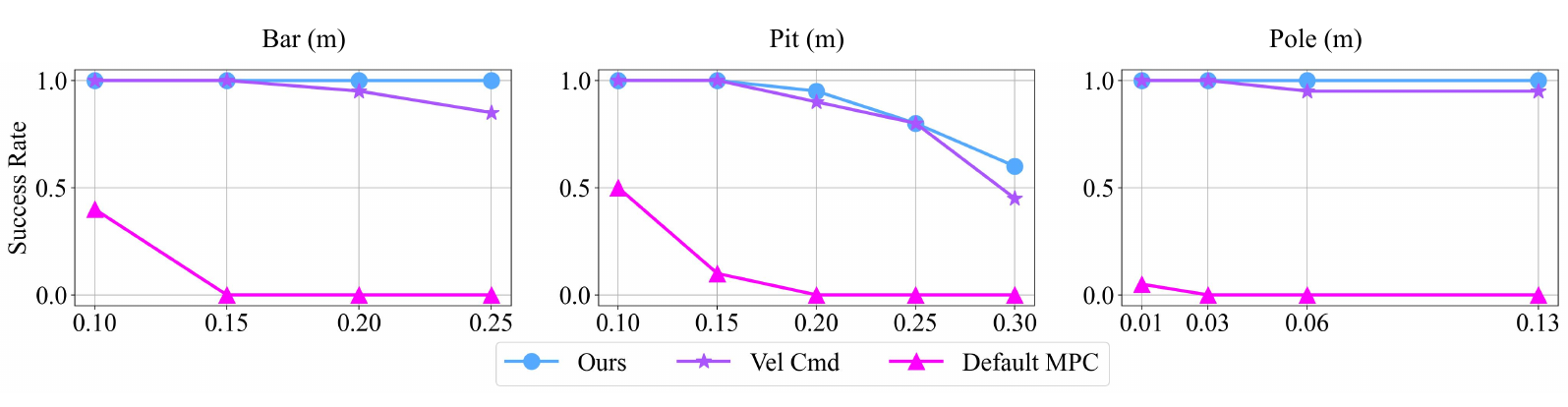}
    \vspace{-4mm}
    \caption{Real-world experiments. The height of the bar is between the range [0.1m, 0.25m]. The width of the pit is between the range [0.1m, 0.3m]. The width of the pole is between the range [0.01m, 0.13m].}
    \label{fig:real_exp}
    \vspace{-6mm}
\end{figure*}

\vspace{-1.5mm}
\subsection{Tiny Trap Benchmark}
\vspace{-0.5mm}
% We design a new tiny trap benchmark in simulation. As shown in Fig.~\ref{fig:benchmark}, we design a runway of 5m$\times$ 60m. The three types of traps are evenly distributed on the runway. There are 10 bars with a height range [0.05m, 0.2m], 50 poles randomly placed in, and 10 pits with a width range [0.05m, 0.2m]. For each experiment, we deployed 1,000 robots. The robots start from the left side of the runway and need to continually pass through all the traps to reach the right side. We call this the ``Mix'' benchmark. There are also ``Bar'', ``Pit'', and ``Pole'' benchmarks. Each has only one specific trap, but the number of traps triples.
We design a new \textbf{Tiny Trap Benchmark} in simulation. As shown in Fig.~\ref{fig:benchmark}, the benchmark consists of a 5m$\times$60m runway with three types of traps evenly distributed along the path. The traps include 10 bars with heights ranging from 0.05m to 0.2m, 50 randomly placed poles, and 10 pits with widths ranging from 0.05m to 0.2m. For each experiment, 1,000 robots are deployed, starting from the left side of the runway and passing through all the traps to reach the right side. We refer to this as the ``Mix'' benchmark. Additionally, there are separate ``Bar,'' ``Pit,'' and ``Pole'' benchmarks, each focusing on one specific type of trap, but with triple the number of traps.

% The benchmark has two metrics: \textbf{Success Rate} and \textbf{Average pass time}. If a robot can reach within 0.2m of the target point within 300 seconds, we record it as successful and record its pass time. The failure cases include falling off the runway, getting stuck halfway, or rolling over on its side. If a robot fails, we record the pass time equal to the max time of 300 seconds. Finally, we calculate the success rate and average pass time. In evaluation, We use fake goal commands same as the real-world deployment, and guide the robot to keep in the center of the runway.
The benchmark uses three metrics: \textbf{Success Rate}, \textbf{Average Pass Time}, and \textbf{Average Travel Distance}. A robot is considered successful if it reaches within 0.2m of the target point within 300 seconds, at which point we record its pass time. Failure cases include falling off the runway, getting stuck, or rolling over. For failed cases, the pass time is set to a maximum of 300 seconds. We then calculate the overall success rate and average pass time. At the end of the evaluation, we average the lateral travel distance of all robots. During the evaluation, we use the same fake goal commands as in real-world deployment to guide the robot to stay in the center.

\vspace{-2mm}
\subsection{Simulation Experiments}
\vspace{-1mm}

We conducted experiments with the following methods: 

\begin{itemize}
\vspace{-0.5mm}
    \item Different combinations of joint links. (Prop: Only trained with proprioception).
    \item Ours w/o goal command: use traditional velocity command, but only train moving forward towards the trap.
    \item Ours w/ Boolean: without encoder, boolean-value collision states are directly feed into low-level RNN, partly like \cite{cheng2024quadruped}. Other settings are same as our method.
    \item RMA~\cite{kumar2021rma}: 
    % A 1D-CNN is used as an adaptation module, employing asynchronously. The teacher-student training framework is used.
    A 1D-CNN serves as an asynchronous adaptation module in the teacher-student training framework.
    \item MoB~\cite{margolis2023walk}: ``Learning a single policy that encodes a structured family of locomotion strategies that solve training tasks in different ways.''
    \item HIMLoco~\cite{long2023him}: ``HIM only explicitly estimates velocity and implicitly simulates the system response as an implicit latent embedding by constrastive learning.'' 
    \vspace{-0.5mm}
\end{itemize}

For RMA and HIMLoco, they are designed to deal with common terrains like stairs or uneven planes. We retrain the policy in our trap terrain, but the rewards and training settings have not changed (e.g. randomly sample velocity). This may cause the relatively poor performance of these two methods.

We report the results in Table. \ref{tab:comparison} and Table. \ref{tab:ablation}. In comparison experiments, our method outperforms others in all metrics. In ablation experiments, policy that utilizes all joint links performs best on ``Mix'' and perform relatively well on other benchmarks. This proves contact force of every joint is helpful in improving the policy performance.

\vspace{-1.5mm}
\subsection{Real-world Experiments}
\label{subsec:real_exp}
\vspace{-0.5mm}

% We deploy the policy to real robots and conduct a series of real-world experiments. Using the teleoperation strategy mentioned in Sec.~\ref{subsec:train_deploy}, we can zero-shot enable the robot to move omnidirectional.
We deploy the policy on real robots and conduct real-world experiments. The robot achieves zero-shot omnidirectional movement using the teleoperation strategy from Sec.~\ref{subsec:train_deploy}.

\begin{figure}[h]
    \centering
    \includegraphics[width=1.0\linewidth]{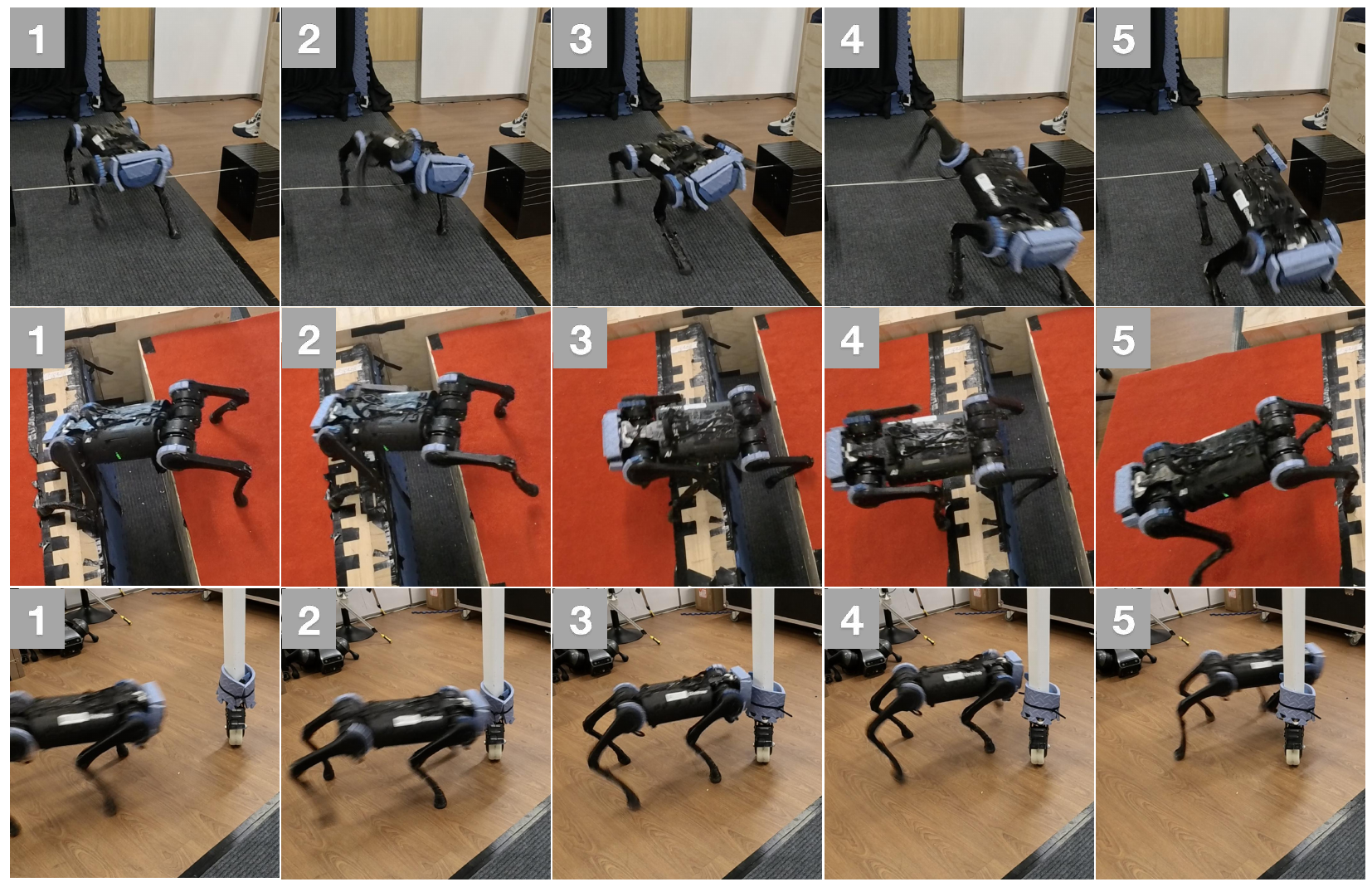}
    \vspace{-6mm}
    \caption{The quadruped robot is passing through tiny traps.}
    \label{fig:real-pic}
    \vspace{-8.5mm}
\end{figure}

% For bar-type traps, our robot has learned to step back its front legs after it touches the obstacle, thus crossing the bar. When the hind legs approach the bar, the robot can predict the imminent collision of the hind legs based on the impact felt by the front legs, and automatically lifts its hind legs to cross the bar. Even when we intentionally trap the robot’s hind legs alone, it can still detect the collision and lift its hind legs to cross the bar.
The visualized results are shown in Fig.~\ref{fig:real-pic}. \textbf{1) Bar:} The robot learns to step back its front legs after contacting the bar. Even when the hind legs are intentionally trapped, the robot can still detect the collision and lift its hind legs across the bar. In addition, the bar is somewhat elastic unlike simulations, which also proves the generalization ability of our policy. \textbf{2) Pit:} The robot learns to support its body with the other three legs when one leg steps into a void, lifting the dangling leg out of the pit. Additionally, our robot has learned to forcefully kick its legs to climb out when multiple legs are stuck in the pit. \textbf{3) Pole:} The robot learns to sidestep to the left or right after colliding with a pole, avoiding the pole by a certain distance before moving forward.
% \vspace{-2mm}

In addition, we deploy and evaluate some other policies for quantitative comparison. For each test, we repeatedly conduct 20 trials and calculate the success rate. The results are in Fig.~\ref{fig:real_exp}. It shows our method obtains the best performance compared to other baselines. The method with velocity command is not much worse. This is because in the training we only sample forward velocity command instead of traditional omnidirectional velocity, which greatly increases the performance.

\begin{figure*}[t]
    \centering
    \includegraphics[width=0.85\textwidth]{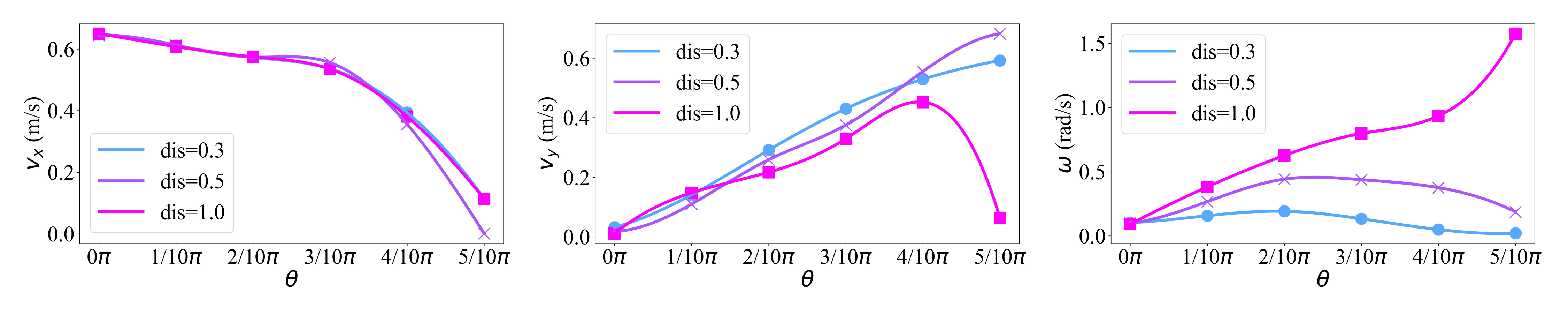}
    \vspace{-4mm}
    \caption{Experiment results of different $\Delta G$. Different goal distances and directions can cause different movements.}
    \label{fig:vel_test}
    \vspace{-6mm}
\end{figure*}

\begin{figure}[h]
    \centering
    \includegraphics[width=1.0\linewidth]{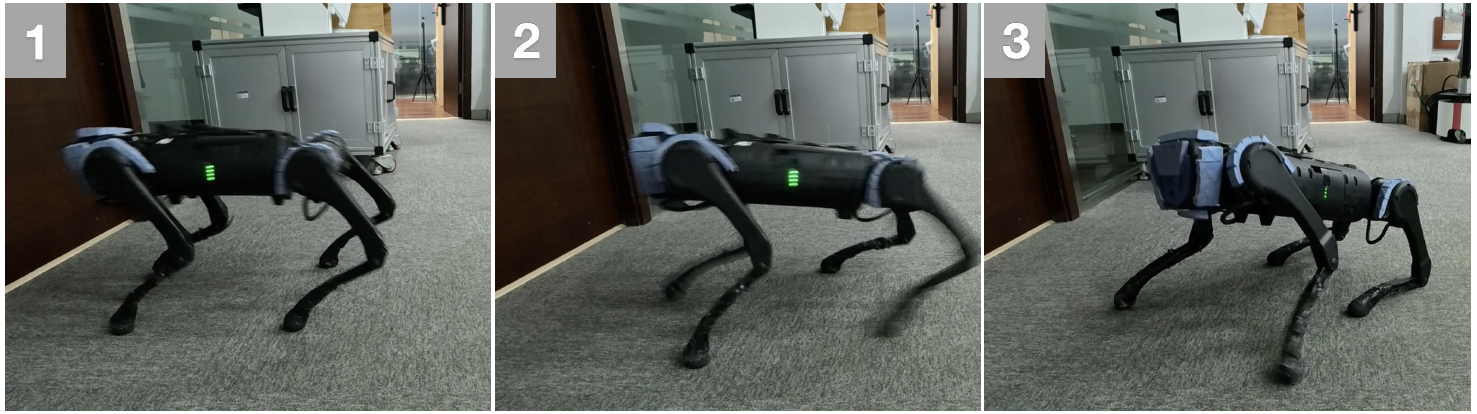}
    \vspace{-5mm}
    \caption{The robot falls down on the ground.}
    \label{fig:boolean}
    \vspace{-7mm}
\end{figure}

We also show that when a boolean value collision state is directly sent to low-level Actor RNN, the robot will face severe sim-to-real gap in deployment. There are ineliminable gaps between sim and real such as friction and motor damping. When the gap occurs, the latent space will have some noise. The latent space of boolean value state is very sparse, as shown in Fig.~\ref{fig:t-sne}, this will cause policy to misjudge the current state more easily and operate irregularly. As shown in Fig.~\ref{fig:boolean}, when the robot moves forward on the plane and suddenly stops for a short time, the gap occurs, and estimated collision state rises to a large number. If we move backward the robot at that time, the robot will assume it has hit a trap. Since there is no trap practically, the robot may touch the ground or even fall down. After introducing contact forces and the contact encoder, the sim-to-real gap is mitigated. The robot can better identify traps and operate more stably.

\vspace{-1.5mm}
\subsection{Additional Experiments}
\label{subsec:additional_exp}
\vspace{-1mm}

We conduct additional experiments to further illustrate our method. All experiments in this section are done in simulation.

\begin{figure}[h]
    \centering
    \includegraphics[width=0.9\linewidth]{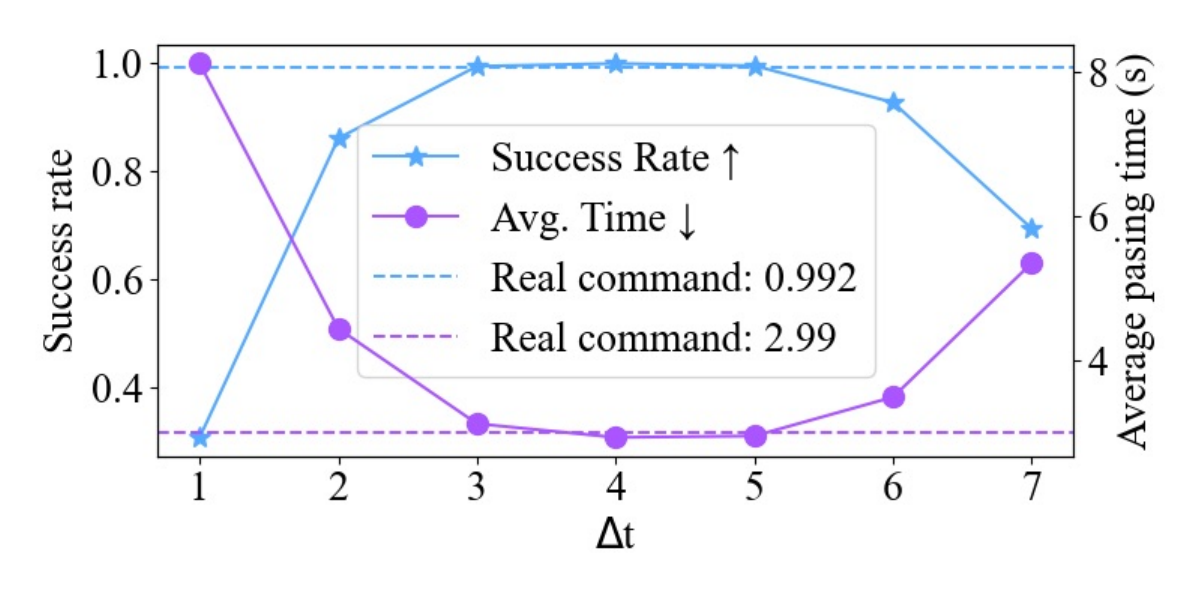}
    \vspace{-5mm}
    \caption{Experiment results of different $\Delta t$. The fake goal command is comparable to or better than the real one.}
    \label{fig:delta_t}
    \vspace{-3mm}
\end{figure}

% \textbf{1) Different $\Delta t$ of the fake goal command.} We conduct a simple test to pass through one bar of 0.2m high. The results are in Fig.~\ref{fig:delta_t}. Unlike previous works\cite{rudin2022advanced}, $\Delta t$ is not an indication of how aggressive the policy is because we have the velocity limit reward. We find the policy with a constant $\Delta t\in[3,5]$ will perform better. It also shows that fake goal command is comparable to or even better than real goal command. However, too large and too small $\Delta t$ can lead to performance degradation. Due to the terrain setting, the robot usually hits traps in the middle of the episode. This makes fewer samples to the large and small $\Delta t$, and worse performance.
\textbf{1) Different $\Delta t$ of the fake goal command.} We tested different $\Delta t$ values for the fake goal command when passing one 0.2m high bar in Fig.~\ref{fig:delta_t}. Unlike previous work\cite{rudin2022advanced}, $\Delta t$ doesn't indicate policy aggressiveness due to the velocity limit reward. A constant $\Delta t\in[3,5]$ yields the best performance, with fake goal commands performing comparable or better than real ones. Too large or small $\Delta t$ values degrade performance, likely due to fewer samples in ranges in the middle of the episode.

\vspace{-4mm}
% \vspace{-2mm}

\textbf{2) Different $\Delta G$ of the fake goal command.} 
% As mentioned in Sec.~\ref{subsec:train_deploy}, different constant goal commands will cause different movements. We record the $V_x$, $V_y$, and $\omega$ for one second and average it. The results are shown in Fig.~\ref{fig:vel_test}, where $\Delta G=(dis\times\cos{\theta}, dis\times\sin{\theta}, 0)$. We can summarize that: $V_x$ is approximately proportional to $\Delta x~(\cos{\theta})$; When \textit{dis} is small, $V_y$ is approximately proportional to $\theta$. When \textit{dis} is large, $V_y$ is still approximately proportional to $\theta$ for small $\theta$, but quickly drops when $\theta$ is close to $\pi/2$; When \textit{dis} is very small, $\omega$ is close to zero. As the \textit{dis} increases, $\omega$ also increases. Theoretically, if we need to conduct a command of $(V_x, V_y, \omega)$, we can firstly approximately calculate $\theta$ based on $V_x:V_y$, and then choose a right \textit{dis} to conduct $\omega$. The limitation of our method is the $\Vert V_{x,y}\Vert_2$ cannot be changed. We can only control velocity direction.
As discussed in Sec.~\ref{subsec:train_deploy}, different constant goal commands result in different movements. We recorded $V_x$, $V_y$, and $\omega$ over one second and averaged the results, as shown in Fig.~\ref{fig:vel_test}, where $\Delta G = (dis \cdot \cos{\theta}, dis \cdot \sin{\theta}, 0)$. The results indicate that: \textbf{a.} $V_x$ is approximately proportional to $\Delta x~(\text{i. e.}~\cos{\theta})$. \textbf{b.} $V_y$ is proportional to $\theta$ with small \textit{dis}; $V_y$ remains proportional to small $\theta$ but drops quickly as $\theta$ approaches $\pi/2$ with large \textit{dis}. \textbf{c.} $\omega$ is near zero but increases with \textit{dis}. In theory, to execute a command of $(V_x, V_y, \omega)$, we can first calculate $\theta$ based on the $V_x:V_y$, then select the appropriate \textit{dis} for $\omega$. The limitation of this approach is that $\Vert V_{x,y}\Vert_2$ remains constant, allowing control over velocity direction only.

\begin{figure}[t]
    \centering
    % \vspace{-1mm}
    \includegraphics[width=\linewidth]{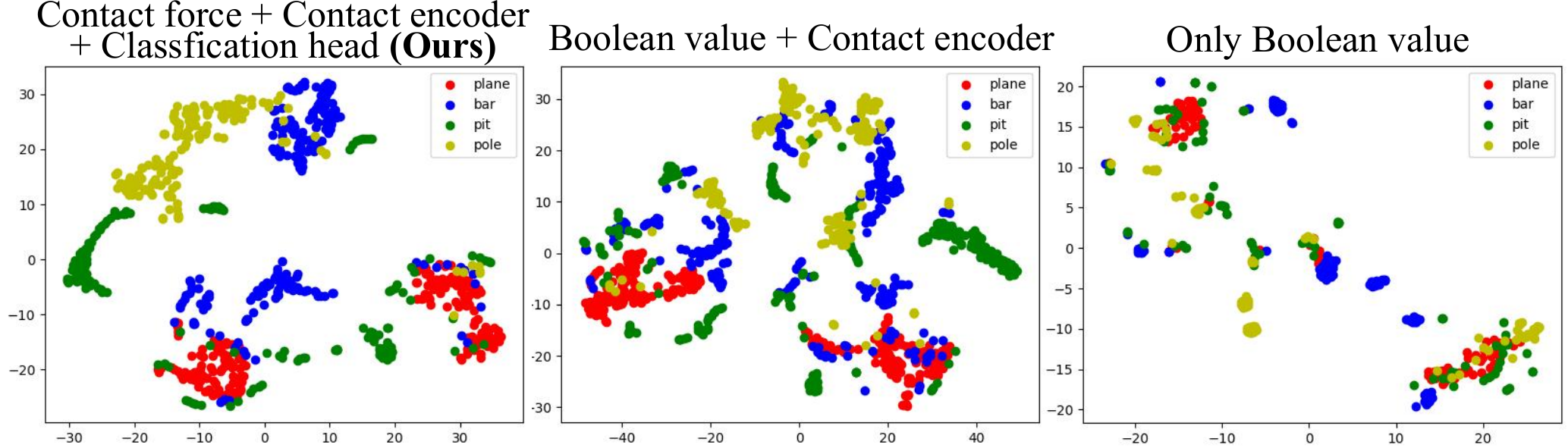}
    \vspace{-5mm}
    % \caption{T-SNE result. The right means using the same contact encoder, but the input is a boolean values collision state without classify loss.}
    \caption{T-SNE results. Middle: using the boolean collision state and an encoder without classification loss. Right: only using boolean collision without encoder and classification loss.}
    \label{fig:t-sne}
    \vspace{-5mm}
\end{figure}

\textbf{3) T-SNE experiments.} We collect the estimated dual-state $\boldsymbol{l}_t$ during the passing time of each trap and visualize them using the t-SNE method\cite{van2008visualizing}. The result is in Fig.~\ref{fig:t-sne}. It shows that our method with classification head and force contact has a more separable and continuous encoding, which means our policy can identify and react to traps more effectively.
% \vspace{-2mm}

\textbf{4) Importance analysis:} 
% Inspired by saliency map analysis\cite{simonyan2013deep} in the computer vision area, we conduct saliency experiments to analyze the importance of each input of state space. First, we obtain the Jacobian matrix $J\in\mathbb{R}^{m\times n}$ by calculating the partial derivative of $m$ output dimensions versus $n$ input dimensions. We sum the absolute values of the output dimensions and normalize them to obtain relative importance vector $I\in\mathbb{R}^{n}$. We collect data for $t$ steps to get $I_t\in\mathbb{R}^{n\times t}$. Each set of different types of inputs is then taken and averaged, proportionally aligned according to their different upper and lower bounds. The results are in Fig.~\ref{fig:errorbar}. In latent space, the contact force takes about 25\% importance against linear velocity and robot friction. In contact force space, the base link is the most important, while others are basically the same. The results prove once again that the estimation of contact forces is very important, while each joint link is not negligible.
Inspired by saliency map analysis~\cite{simonyan2013deep}, we conducted experiments to analyze the importance of each input in the state space. By calculating the Jacobian matrix and normalizing the results, we found that contact force accounts for about 25\% of the importance relative to linear velocity and robot friction. In the contact force space, the base link is the most critical, while the other links show similar importance. These results highlight the crucial role of contact force estimation and the significance of each joint link. Formulas and details can be found in \textit{Appendix F}.

\vspace{-4mm}
\begin{figure}[h]
    \centering
    \includegraphics[width=0.95\linewidth]{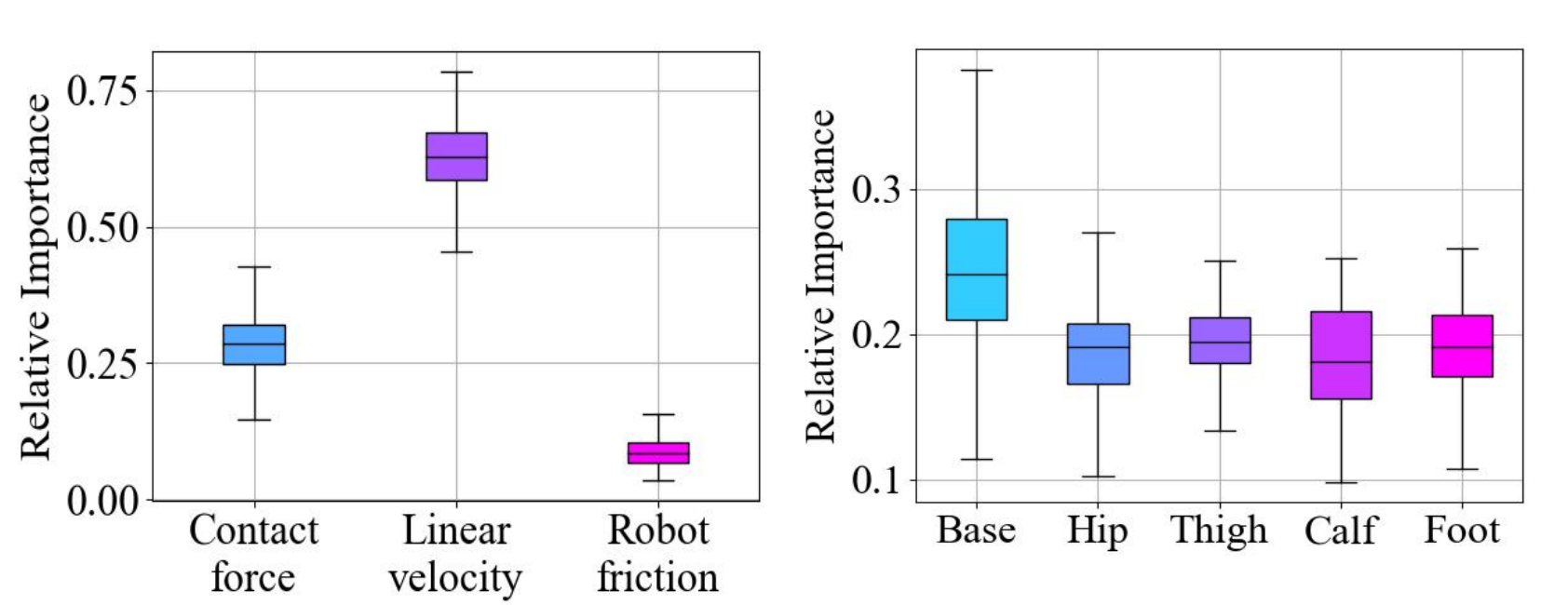}
    \vspace{-1mm}
    \caption{Importance analysis on latent space and contact links.}
    \label{fig:errorbar}
    \vspace{-5mm}
\end{figure}

\section{CONCLUSION}
In this work, we propose a novel method for enabling a quadruped robot to pass through tiny traps using only proprioception, avoiding the imprecision of camera images. A contact encoder and classification head are used to capture contact forces latent. With well-designed rewards and fake goal commands, we enable approximate omnidirectional movement without localization techniques. We deploy our policy on the real robot and demonstrate robustness through extensive experiments.

% However, there are some limitations. We can only control the direction of linear velocity, not its magnitude, due to our dense reward settings with a velocity limit. Future work could consider taking velocity magnitude as an input. Additionally, our policy handles hard or slightly bouncy traps but struggles with traps causing large deformations, such as rubber bands. Real-world fine-tuning may help address this issue.
However, limitations include the inability to control velocity magnitude due to reward settings and struggles with highly deformable traps like rubber bands. Future work could address these with velocity adjustments and real-world fine-tuning.

% \addtolength{\textheight}{-12cm}   % This command serves to balance the column lengths
                                  % on the last page of the document manually. It shortens
                                  % the textheight of the last page by a suitable amount.
                                  % This command does not take effect until the next page
                                  % so it should come on the page before the last. Make
                                  % sure that you do not shorten the textheight too much.

%%%%%%%%%%%%%%%%%%%%%%%%%%%%%%%%%%%%%%%%%%%%%%%%%%%%%%%%%%%%%%%%%%%%%%%%%%%%%%%%

%%%%%%%%%%%%%%%%%%%%%%%%%%%%%%%%%%%%%%%%%%%%%%%%%%%%%%%%%%%%%%%%%%%%%%%%%%%%%%%%

\clearpage
\newpage
%%%%%%%%%%%%%%%%%%%%%%%%%%%%%%%%%%%%%%%%%%%%%%%%%%%%%%%%%%%%%%%%%%%%%%%%%%%%%%%%
\section*{APPENDIX}

In \textit{Appendix}, we further illustrate the technical details of training, deploying, and experiment setting. \textit{Appendix} mainly has these sections:
\begin{itemize}
    \item \textbf{A. Reward Functions: }the formula of each reward, the definition and analysis of regularization reward, and style reward.
    \item \textbf{B. Network Architectures: }the network architectures of Actor RNN, Critic RNN, Estimator Module, Latent Encoder, and Contact Encoder.
    \item \textbf{C. Dynamic Randomization: }details of the domain randomization and Gaussian noise.
    \item \textbf{D. Trap Terrain Setting in Simulation: }the trap terrain details for training.
    \item \textbf{E. Training Hyperparameters: }the hyperparameters of PPO, contact force, and t-SNE visualization.
    \item \textbf{F. Importance Analysis: }the method and formula for importance analysis.
    \item \textbf{G. Real-world Experiment Settings and Additional Results:} details of the experimental equipment and deployment, and additional experiments in the low-light environment.
\end{itemize}

\begin{table*}[bp]
    \renewcommand{\arraystretch}{1.3}
    \centering
    \caption{Reward functions}
    \label{tab:reward}
    \setlength\tabcolsep{14pt}
    % \fontsize{10}{12}\selectfont
    \begin{tabular}{c|c|c|c}
        \toprule
         \textbf{Type} & \textbf{Item} & \textbf{Formula} & \textbf{Weight}\\ \midrule
         \multirow{5}{*}[-7mm]{\textbf{Task}} & Get goal & $\displaystyle \frac{1}{0.4+\Vert \Delta G\Vert_2}$ & 5.0\\ 
        & Heading & $\displaystyle \left \{ \begin{matrix}
    \vspace{2mm}
    \displaystyle \frac{\Delta x}{\Vert \Delta G\Vert_2+\epsilon}, &\Vert \Delta G\Vert_2 \neq 0\\
    1,&\Vert \Delta G\Vert_2 = 0\\
    \end{matrix} \right.$ & $3.0$ \\
    & Finish vel & $(\displaystyle \Vert v\Vert+\Vert \omega\Vert)\cdot(\Vert \Delta G\Vert_2<0.2)$ & -1.0\\
    & Finish pos & $\displaystyle (\sum^{12}_{i=1}{\lvert q-q_{default}\rvert)\cdot(\Vert \Delta G\Vert_2<0.2)}$ & -1.0\\
     & Alive & 1 & $3.0$ \\ \midrule
        \multirow{7}{*}{\textbf{Regularization}} 
        & Stall & $(\Vert V_{x,y}\Vert_2<0.1)\cdot(\Vert \Delta G\Vert_2>0.25)$ & $-2.0$ \\
        
        & Vel limit & $(\omega_{z}<\omega_{limit})\cdot(\Vert v_{x,y}\Vert_2<v_{limit})$ & $2.0$ \\
        & Joint vel & $-\|\dot{\mathbf{q}}\|_2$ & $0.002$ \\
        & Joint acc & $-\|\ddot{\mathbf{q}}\|_2$ & $2\times10^{-6}$ \\
        & Ang vel stability & $-(\|\omega_{t,x}\|_2+\|\omega_{t,y}\|_2)$ & $0.2$ \\
         & Feet in air & $\displaystyle\sum_{i=0}^{3}\left(\mathbf{t}_{air,i}-0.3\right)+10\cdot\max\left(0.5-\mathbf{t}_{air,i},0\right)$ & $0.05$ \\ 
        & Balance & $\| F_{feet,0}+F_{feet,2}-F_{feet,1}-F_{feet,3} \|_2$ & $-2\times10^{-5}$ \\ \midrule
    \textbf{Style} & AMP & $\max\left[0,1-0.25\left(\mathcal{D}_{amp}(s_t,s_{t+1})-1\right)^2\right]$ & 0.1 \\
    \bottomrule
    \end{tabular}
\end{table*}

\subsection{Reward Functions}
\label{subsec:reward_appendix}

The reward function has three components: task reward $r_t^T$, regularization reward $r_t^R$, and style reward $r_t^S$. 

In Sec.~\ref{subsec:reward}, we have already introduced the \textbf{task reward}, which plays a major role in the training. In addition, \textbf{regularization reward} is used to optimize the performance of the robot. ``Stall'' reward is used to prevent the robot from stopping in the middle of the journey. ``Velocity limit'' reward is used to slow down the robot and ensure safety. ``Joint velocity'' reward and ``Joint acceleration'' reward are used to make the joint movements more stable and smooth. ``Angular velocity stability'' reward is used to make the base of the robot more stable. ``Feet in air'' reward is used to improve the gait and prevent the feet from rubbing on the ground. ``Balance reward'' is used to improve the left-right symmetry. For \textbf{style reward}, we first collect a dataset using an MPC controller. The dataset contains a state transition $(s_t, s_{t+1})$, with a time interval same as the RL policy. $s_t\in\mathbb{R}^{19}$ includes joint positions, base height, base linear velocity, and base angular velocity. We randomly select 200 velocity commands in the simulator. Each command lasts for two seconds and is converted to the next command continually. Following \cite{peng2021amp, wu2023learning}, we train a Discriminator $\mathcal{D}_{amp}$ by the following loss function:

\vspace{-4mm}
% \begin{equation}
% \begin{gathered}
\begin{align*}
L_{discriminator}&=\mathbb{E}_{(s_{t},s_{t+1})\sim \text{MPC}}\left[\left(\mathcal{D}_{amp}\left(s_{t},s_{t+1}\right)-1\right)^{2}\right] \\
&+\mathbb{E}_{(s_t,s_{t+1})\sim\text{Policy}}\left[\left(\mathcal{D}_{amp}\left(s_t,s_{t+1}\right)+1\right)^2\right] \\
&+\alpha^{gp}\mathbb{E}_{(s_t,s_{t+1})\sim\text{MPC}}\begin{bmatrix}\left\|\nabla \mathcal{D}_{amp}\left(s_t,s_{t+1}\right)\right\|_2\end{bmatrix}, 
% \end{gathered}
% \end{equation}
\end{align*}

And then we use $\mathcal{D}_{amp}$ to score the gait performance from policy output $(s_t,s_{t+1})$:
\vspace{-1mm}
\begin{equation}
    r_{style}=\max\left[0,1-0.25\left(\mathcal{D}_{amp}(s_t,s_{t+1})-1\right)^2\right].
\end{equation}

\vspace{-2mm}

\subsection{Network Architectures}
\label{subsec:network}

The details of network architectures are shown in Tab.~\ref{tab:network}.
\begin{table}[htbp]
    \vspace{-1mm}
    \renewcommand{\arraystretch}{1.3}
    \caption{\label{tab:network}Network architecture details}
    \vspace{-1mm}
    \centering
    \begin{tabular}{c|cccc}
        \toprule
        \textbf{Network} & \textbf{Type} & \textbf{Input} & \textbf{Hidden layers} & \textbf{Output}\\ \midrule
        Actor RNN & LSTM & $\boldsymbol{p}_t$, $\boldsymbol{l}_t$, $\boldsymbol{g}_t$ & [512, 256] & $\boldsymbol{a}_t$\\
        Critic RNN & LSTM & $\boldsymbol{p}_t$, $\hat{\boldsymbol{s}}_t$, $\hat{\boldsymbol{c}_t}$, $\boldsymbol{g}_t$ & [512, 256] & $\boldsymbol{V}_t$\\
        Estimator Module & LSTM & $\boldsymbol{p}_t$, $\boldsymbol{g}_t$ & [256, 256] & $\boldsymbol{h}_t$\\
        Latent Encoder & MLP & $\boldsymbol{h}_t$ & [256, 256] & $\boldsymbol{l}_t$\\
        Contact Encoder & MLP & $\hat{\boldsymbol{c}_t}$ & [32, 16] & $\boldsymbol{l}_{t_c}$\\
        \bottomrule
    \end{tabular}
    \vspace{-2mm}
\end{table}

\subsection{Dynamic Randomization}

For better sim-to-real transfer, we introduce dynamic randomization including domain randomization and Gaussian noise. We have a series of domain randomizations including base mass, mass position, friction, initial joint positions, initial base velocity, and motor strength. the random ranges are shown in the Tab.~\ref{tab:random}.

\begin{table}[htbp]
    \vspace{-1mm}
    \renewcommand{\arraystretch}{1.3}
    \centering
    \caption{\label{tab:random}Domain randomization}
    \vspace{-1mm}
    \begin{tabular}{c|cc}
        \toprule
        \textbf{Parameters} & \textbf{Range} & \textbf{Unit} \\ \midrule
        Base mass & [0, 3] & $kg$\\
        Mass position of X axis & [-0.2, 0.2] & $m$\\
        Mass position of Y axis & [-0.1, 0.1] & $m$\\
        Mass position of Z axis & [-0.05, 0.05] & $m$\\
        Friction & [0, 2] & - \\
        Initial joint positions & [0.5, 1.5]$~\times~$nominal value & $rad$\\
        Initial base velocity & [-1.0, 1.0] (all directions) & $m/s$\\
        Motor strength & [0.9, 1.1]$~\times~$nominal value & - \\
        Proprioception latency & [0.2, 0.4] & $s$\\
        \bottomrule
    \end{tabular}
    \vspace{-2mm}
\end{table}

Besides, we add Gaussian noise to the input observation, as shown in Tab.~\ref{tab:gaussian}. This aims to simulate the noise of real robot sensors. Lots of experiments show that with dynamic randomization, the policy can be easily transferred from simulation to the real world without additional training.

\begin{table}[htbp]
    \vspace{-1mm}
    \renewcommand{\arraystretch}{1.3}
    \centering
    \caption{\label{tab:gaussian}Gaussian noise}
    \vspace{-1mm}
    \begin{tabular}{c|cc}
        \toprule
        \textbf{Observation} & \textbf{Gaussian Noise Amplitude} & \textbf{Unit} \\ \midrule
        Linear velocity & 0.05 & $m/s$\\
        Angular velocity & 0.2 & $rad/s$\\
        Gravity & 0.05 & $m/s^2$\\
        Joint position & 0.01 & $rad$\\
        Joint velocity & 1.5 & $rad/s$\\
        \bottomrule
    \end{tabular}
    \vspace{-4mm}
\end{table}

\subsection{Trap Terrain Setting in Simulation}

\begin{figure}[h]
    \vspace{-4mm}
    \centering
    \includegraphics[width=1.0\linewidth]{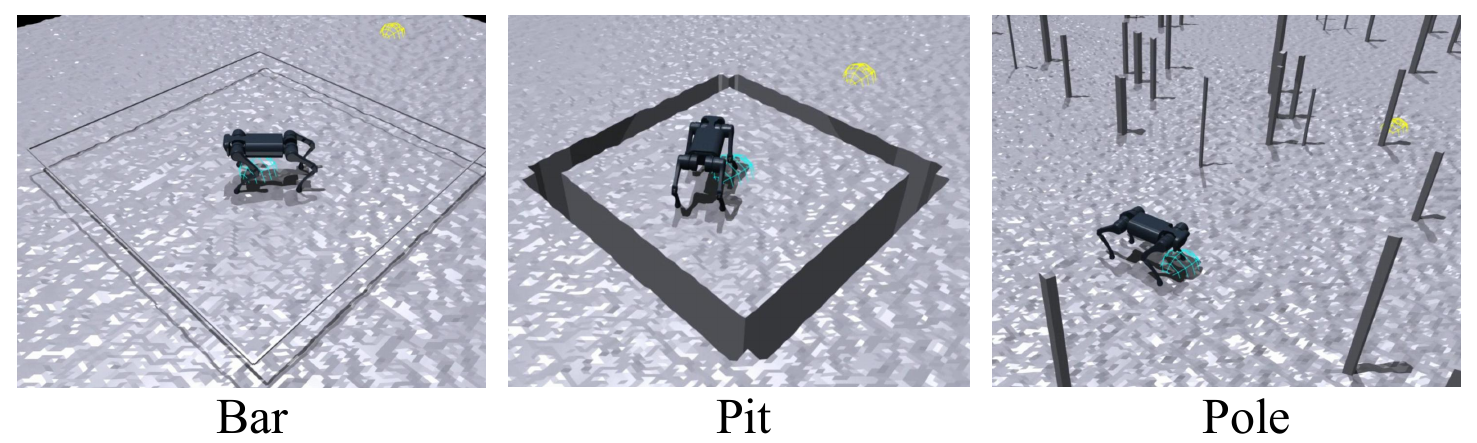}
    \vspace{-5mm}
    \caption{Trap Terrain setting.}
    \label{fig:trap_setting}
    \vspace{-3mm}
\end{figure}

We employ ``Terrain Curriculum'' introduced in previous work\cite{heess2017emergence} for better policy training. Due to the instability of reinforcement learning in early training, it is difficult for the policy to learn the movement in complex traps at once. Therefore, we design a trap curriculum to guide the policy from easy to difficult. The terrain is distributed in 10 rows and 10 columns. The terrains are divided into $4$ categories. Each categorey has different traps ranging from easy to difficult, consisting of $10$ variations. The column numbers of Bar, Pit, Pole, and Plane are 3,2,3,2. To prevent the robot from cheating by detouring, we put the bar and pit in a circle. As shown in Fig.~\ref{fig:trap_setting}, the robot is born inside the circle (blue point) and needs to reach outside the circle (yellow point). The height of the bar increases evenly from 0.05m to 0.25m, with a width randomizing in the range [0.025m, 0.1m]. The width of the pit increases evenly from 0.05m to 0.30m. The number of the pole increases evenly from 10 to 60, with a width randomizing in the range [0.025m, 0.1m]. In addition, we add perlin noise to all of the terrains with an amplitude in the range [0.05m, 0.15m].

\subsection{Training Hyperparameters}

In our work, we conduct a Policy Optimization algorithm (PPO) as our reinforcement learning method. The hyperparameters are shown in Tab.~\ref{tab:ppohyper}.

\begin{table}[htbp]
    \vspace{-1mm}
    \renewcommand{\arraystretch}{1.3}
    \setlength\tabcolsep{18pt}
    \caption{PPO hyperparameters}
    \vspace{-1mm}
    \centering
    \label{tab:ppohyper}
    \begin{tabular}{c|c}
        \toprule
        \textbf{Hyperparameter} & \textbf{Value}  \\ \midrule
        clip min std & 0.05 \\
        clip param & 0.2\\
        desired kl & 0.01\\
        entropy coef & 0.01\\
        gamma & 0.99\\
        lam & 0.95\\
        learning rate & 0.001\\
        max grad norm & 1\\
        num mini batch & 4\\
        num steps per env & 24\\
        \bottomrule
    \end{tabular}
    \vspace{-2mm}
\end{table}

In the training step, we clip the contact force to [0$N$, 100$N$].

In addition, we conduct t-SNE visualization in our additional experiments. The hyperparameters are shown in Tab.~\ref{tab:tsne}.

\begin{table}[htbp]
    \vspace{-1mm}
    \renewcommand{\arraystretch}{1.3}
    \setlength\tabcolsep{18pt}
    \caption{\label{tab:tsne}T-SNE hyperparameters}
    \vspace{-1mm}
    \centering
    \begin{tabular}{c|c}
        \toprule
        \textbf{Hyperparameter} & \textbf{Value}  \\ \midrule
        init & `random' \\
        perplexity & 30\\
        learning rate & 200\\
        \bottomrule
    \end{tabular}
    \vspace{-4mm}
\end{table}

\subsection{Importance Analysis}

Assume the input $I\in \mathbb{R}^N$ and the output (action) $O\in \mathbb{R}^M$.
\begin{equation}
    O=\text{Policy}(I),
\end{equation}

First, we obtain the Jacobian matrix $J\in \mathbb{R}^{M\times N}$ by calculating the partial derivative.

\begin{equation}
J = 
	\begin{bmatrix}
		\vspace{1.5ex}
		\dfrac{\partial O_1}{\partial I_1} & \dfrac{\partial O_1}{\partial I_2}  & \cdots & \dfrac{\partial O_1}{\partial I_n}\\ 
		\vspace{1.5ex}
		\dfrac{\partial O_2}{\partial I_1} & \dfrac{\partial O_2}{\partial I_2}  & \cdots & \dfrac{\partial O_2}{\partial I_n}\\
		\vspace{1.5ex}
		\vdots                             & \vdots                              & \vdots & \vdots                            \\
		\vspace{1.5ex}
		\dfrac{\partial O_m}{\partial I_1} & \dfrac{\partial O_m}{\partial I_2}  & \cdots & \dfrac{\partial O_m}{\partial I_n}
	\end{bmatrix},
\end{equation}

We take the absolute value for each term of the matrix.
\vspace{-1mm}
\begin{equation}
J_{abs} = \lvert J\rvert,
\end{equation}

For each input $I_i$, we sum the corresponding output dimensions and align them with the upper $U_i$ and lower bounds $L_i$ to get the importance vector $S\in \mathbb{R}^{N}$.
\vspace{-1mm}
\begin{equation}
    S_i=(U_i-L_i)\cdot \sum_{j=1}^{M} J_{abs}(j,i),~~i=1,2,\cdots, N
\end{equation}

For a group of input $\mathcal{G}\in I$, such as Contact force, Linear velocity, etc, we average importance for every input in $\mathcal{G}$.
\vspace{-1mm}
\begin{equation}
    S_{\mathcal{G}}=\frac{\sum_{I_i\in\mathcal{G}} S_i}{\text{num}(I_i\in \mathcal{G})},
\end{equation}

The group $\mathcal{G}_1, \mathcal{G}_2, \cdots, \mathcal{G}_k$ is for comparison, we normalize them to get relative importance $R\in \mathbb{R}^{k}$ for each group.
\vspace{-1mm}
\begin{equation}
    R_i=\frac{S_{\mathcal{G}_i}}{\sum_{j=1}^{N} S_{\mathcal{G}_j}},~~i=1,2,\cdots, k
\end{equation}

\subsection{Real-world Experiment Settings and Additional Results}
\vspace{-1mm}

\begin{figure*}[t]
    \centering
    \includegraphics[width=0.94\linewidth]{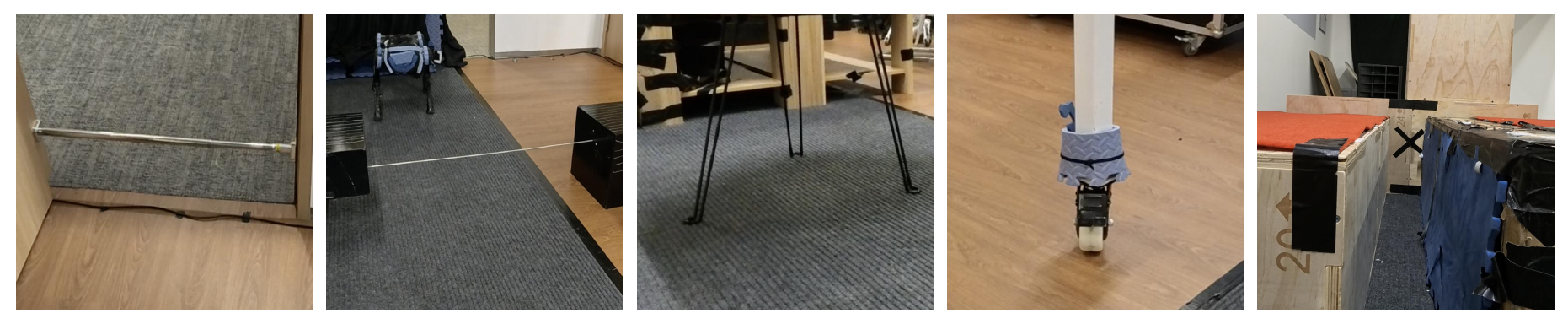}
    \vspace{-1mm}
    \caption{Real-world traps: Thick Bar, Thin Bar, Thin Pole, Thick Pole, Pit.}
    \label{fig:real-obstacles}
    \vspace{-1mm}
\end{figure*}

\begin{figure*}[t]
    \vspace{-2mm}
    \centering
    \includegraphics[width=0.94\linewidth]{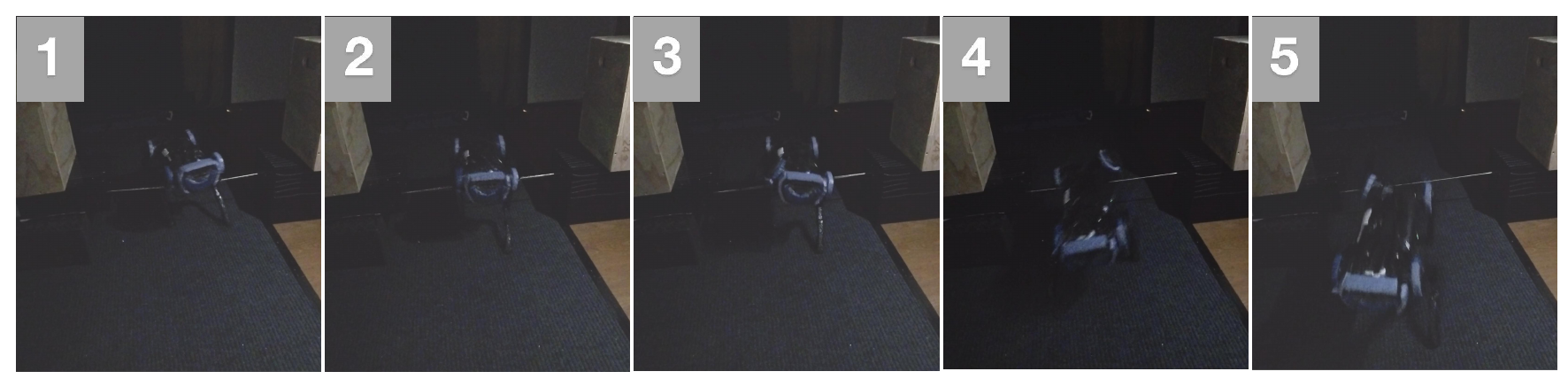}
    \vspace{-1mm}
    \caption{Crossing \textit{Bar} in low-light environment.}
    \label{fig:night-bar}
    \vspace{-1mm}
\end{figure*}

\begin{figure*}[t]
    \vspace{-3mm}
    \centering
    \includegraphics[width=0.94\linewidth]{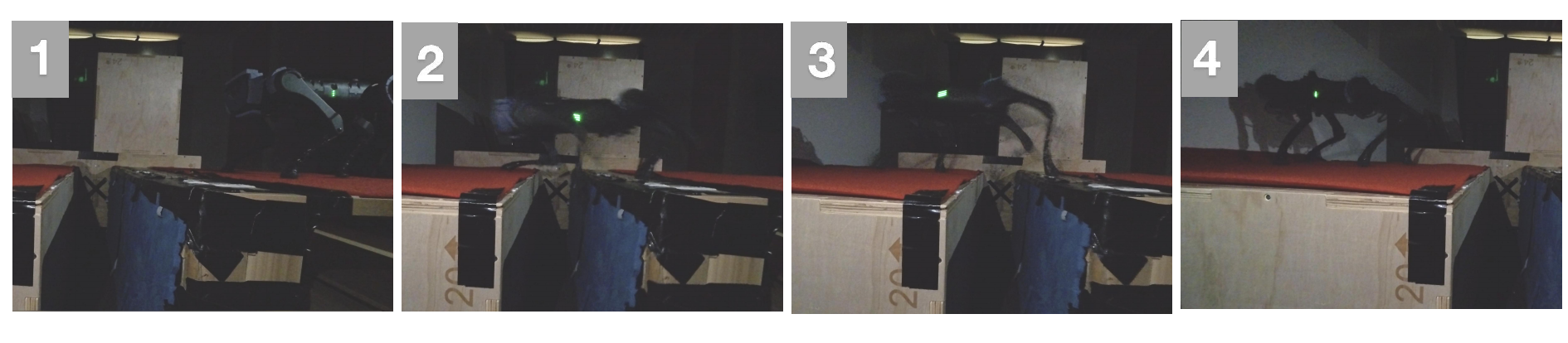}
    \vspace{-3mm}
    \caption{Crossing \textit{Pit} in low-light environment.}
    \label{fig:night-pit}
\end{figure*}

\begin{figure*}[t]
    \vspace{-3mm}
    \centering
    \includegraphics[width=0.94\linewidth]{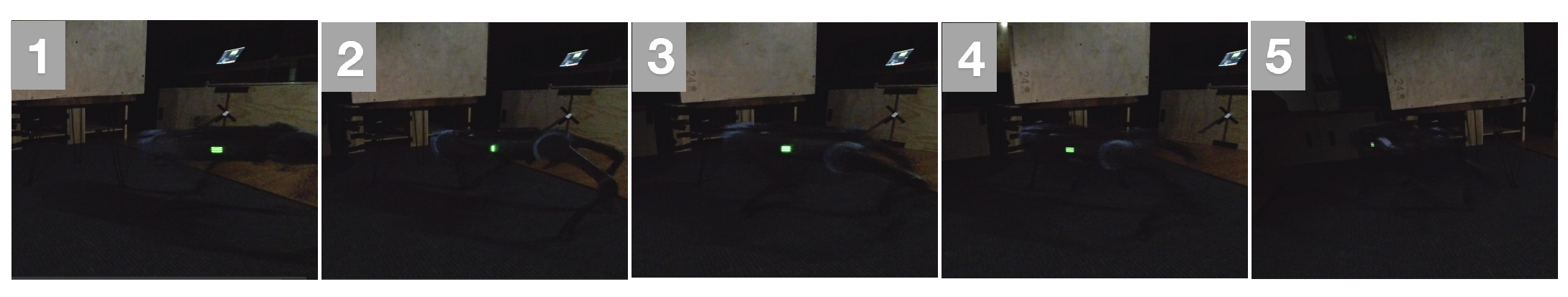}
    \vspace{-1mm}
    \caption{Crossing \textit{Pole} in low-light environment.}
    \label{fig:night-pole}
    \vspace{-7mm}
\end{figure*}

We use common easily accessible items as traps for our real-world experiments, as shown in Fig. \ref{fig:real-obstacles}. For the \textit{Bar} trap, there are two variations: thin bars and thick bars. The thin bars have a diameter of 6mm, while the thick bars measure 20mm in diameter. For the \textit{Pit} trap, we separate two wooden boxes with a height of 40mm by some distance. For the \textit{Pole} trap, there are also thin and thick poles. The thin poles are the legs of a iron table with a diameter of 8mm, while the thick poles include a range of obstacles made from thick iron poles and sticks of varying diameters.

We also conduct experiments in low-light environment, as shown in Fig.~\ref{fig:night-bar}, Fig.~\ref{fig:night-pit}, and Fig.~\ref{fig:night-pole}. The robot can robustly move through different traps even if there is little light. The results demonstrate the effectiveness and importance of proprioception locomotion in scenarios where there is no visual input, such as during nighttime.

%%%%%%%%%%%%%%%%%%%%%%%%%%%%%%%%%%%%%%%%%%%%%%%%%%%%%%%%%%%%%%%%%%%%%%%%%%%%%%%%
\newpage
% \addtolength{\textheight}{-5cm} 
\bibliographystyle{./IEEEtran} % use IEEEtran.bst style
\bibliography{./IEEEabrv,./IEEEexample}

\end{document}